\documentclass[11pt,a4paper]{article}
\usepackage[hyperref]{eacl2021}
\usepackage{times}
\usepackage{latexsym}
\usepackage{amsfonts}
\usepackage{xcolor}

\usepackage{graphicx}
\usepackage{subfigure}
\usepackage{tipa}
\usepackage{listings}

\definecolor{codegreen}{rgb}{0,0.6,0}
\definecolor{codegray}{rgb}{0.5,0.5,0.5}
\definecolor{codepurple}{rgb}{0.58,0,0.82}
\definecolor{backcolour}{rgb}{0.95,0.95,0.92}

\lstdefinestyle{mystyle}{
    backgroundcolor=\color{backcolour},   
    commentstyle=\color{codegreen},
    keywordstyle=\color{magenta},
    numberstyle=\tiny\color{codegray},
    stringstyle=\color{codepurple},
    basicstyle=\ttfamily\footnotesize,
    breakatwhitespace=false,         
    breaklines=true,                 
    captionpos=b,                    
    keepspaces=true,                 
    numbers=left,                    
    numbersep=5pt,                  
    showspaces=false,                
    showstringspaces=false,
    showtabs=false,                  
    tabsize=2
}

\usepackage{hyperref}
\hypersetup{
    colorlinks=true,
    linkcolor=blue,
    filecolor=magenta,      
    urlcolor=cyan,
}

\usepackage{microtype}
\usepackage{verbatim}

\aclfinalcopy 

\title{English-Twi Parallel Corpus for Machine Translation}
\author{Paul Azunre$^{1*}$, Salomey Osei$^{2*}$, Salomey Afua Addo$^{3*}$,  Lawrence Asamoah Adu-Gyamfi$^{*}$, \\
\textbf{Stephen Moore$^{4*}$, Bernard Adabankah$^{5*}$, Bernard Opoku$^{6*}$, Clara Asare-Nyarko$^{4*}$} \\
\textbf{Samuel Nyarko$^{15*}$, Cynthia Amoaba$^{*}$, Esther Dansoa Appiah$^{8*}$, Felix Akwerh$^{2*}$} \\
\textbf{Richard Nii Lante Lawson$^{9*}$, Joel Budu$^{10*}$, Emmanuel Debrah$^{4*}$, Nana Boateng$^{1*}$} \\
\textbf{Wisdom Ofori$^{*}$, Edwin Buabeng-Munkoh$^{*}$, Franklin Adjei$^{11*}$, Isaac Kojo Essel Ampomah$^{12*}$} \\
\textbf{Joseph Otoo$^{13*}$, Reindorf Borkor$^{2*}$, Standylove Birago Mensah$^{2*}$, Lucien Mensah$^{7*}$} \\
\textbf{Mark Amoako Marcel$^{*}$, Anokye Acheampong Amponsah$^{14*}$, and James Ben Hayfron-Acquah$^{2*}.$} \\ \\

 $^*$ NLP Ghana,$^1$ Algorine,$^2$ Kwame Nkrumah University of Science and Technology, \\
 $^3$ Leuphana University Luneburg,$^4$University of Cape Coast, 
 $^5$ Edinburgh Napier University, \\ $^6$ Accra Institute of Technology,$^7$ Tulane University, $^{8}$ University of Tromso, $^{9}$ AiMlCamp, \\ $^{10}$ University of Strathclyde, $^{11}$ Azubi Africa, $^{12}$ Ulster University, \\
 $^{13}$ Centre for Research, Data Science and IT Solutions, $^{14}$ University of Energy and Natural Resources, \\ $^{15}$ Integrated Geospatial Intelligence Application Centre, SRH Berlin University of Applied Science.
  }

\date{}

\begin{document}
\maketitle
\begin{abstract}
We present a parallel machine translation training corpus for English and Akuapem Twi of 25,421 sentence pairs. We used a transformer-based translator to generate initial translations in Akuapem Twi, which were later verified and corrected where necessary by native speakers to eliminate any occurrence of translationese. In addition, 697 higher quality crowd-sourced sentences are provided for use as an evaluation set for downstream Natural Language Processing (NLP) tasks. The typical use case for the larger human-verified dataset is for further training of machine translation models in Akuapem Twi. The higher quality 697 crowd-sourced dataset is recommended as a testing dataset for machine translation of English to Twi and Twi to English models. Furthermore, the Twi part of the crowd-sourced data may also be used for other tasks, such as representation learning, classification, etc. We fine-tune the transformer translation model on the training corpus and report benchmarks on the crowd-sourced test set.
\end{abstract}

\section{Introduction}
The state of Natural Language Processing (NLP) has been undergoing a revolution. Due to the vast increase in the availability of digitised text data, computers are increasingly needed to keep track of signals online, on social media, on the Internet of Things (IoT) and/or in conversations with customers, partners and other stakeholders. The amount of data available is significantly beyond what human analysts can reasonably digest manually. This touches every conceivable application area — from medical diagnosis and financial system interfaces to translation systems and cybersecurity.
Specifically, NLP has made dramatic strides in allowing engineers to reuse NLP knowledge acquired at major laboratories and institutions — such as Google, Facebook or Microsoft — and adapting it to the engineer’s problem very quickly on a laptop or even smartphone. This is loosely called \textit{transfer learning} \citep{rosenstein2005transfer} by the machine learning community. 

Unfortunately, the state of NLP on Ghanaian languages has not kept up with these developments. To date, there is no reliable machine translation system for any Ghanaian languages. This makes it harder for the global Ghanaian diaspora to learn their own languages, something many want to do. It risks their language and culture not being preserved in an increasingly digitised future. It also means that service providers and health workers trying to reach remote areas hit by emergencies, disasters, etc. face needless additional obstacles to providing life-saving care.

To begin closing this gap, in this paper we present a parallel machine translation training corpus for English and Akuapem Twi of 25,421 sentence pairs. We used a transformer-based translator to generate initial translations in Akuapem Twi, which were later verified and corrected where necessary by native speakers. The typical use case for the dataset is for further training of machine translation models or the fine-tuning of an unsupervised embedding for the Akuapem Twi dialect of Akan. Moreover, the data set can also be used for other downstream NLP tasks, such as named entity
recognition and part-of-speech tagging, with appropriate additional annotations.

Furthermore, we present 697 crowd-sourced parallel sentences from native speakers of Twi across the country, collected via Google Forms. This dataset is recommended as a testing dataset for machine translation of English to Twi and Twi to English models.

We fine-tune our transformer translation model on the training corpus and report benchmarks on the crowd-sourced test set.

\section{Related Work}
\subsection{Machine Translation}
Machine translation (MT) converts sentences from a source language to a targeted language using computer systems, with or without human aid \citep{Hutchins1992AnIT}. Historically, MT spans from the 1949 work of Warren Weaver through recent developments like automatic translation from Google \citep{Chragui2012TheoreticalOO}.

In this study, we built on the OPUS-MT English to Twi model \citep{TiedemannThottingal:EAMT2020}. OPUS-MT models are based on state-of-the art transformer-based neural machine translation. The neural network architecture for these models is based on a standard transformer setup with 6 self-attentive layers in both the encoder and decoder network, and 8 attention heads in each layer \citep{vaswani2017attention}. For many low-resource languages, these models can be a good starting point for building useful models. 

In this work, we use the English to Twi OPUS-MT model to generate preliminary translations that are corrected by native speakers. While this approach has some risks -- such as \textit{translationese} or overly literal translations making it into the dataset -- it reduced the cost of generating the data by a factor of approximately 50 percent. This made the project feasible to execute, given the resources and budget available. The resulting data is used to fine-tune the model for better performance. 

\subsection{Description of Akuapem Twi}

Ghana is a multilingual country with at least 75 local languages. Of these, the Twi language of the Akan people is the most widely spoken.
It is spoken as a native language in parts of southern and central Ghana, as well as some areas of Cote d’Ivoire. By some estimates it has approximately 20 million native speakers \citep{Ethniclinguisticgroups}. Akan comprises at least four distinct dialects, namely Asante Twi, Akuapem Twi, Fante and Bono. Knowing the Akan language alone allows you to navigate your way through most parts of Ghana. You are likely to find someone who at the very least understands the language in almost every part of the country.

The description of Akuapem Twi focuses on the syntax and the phonology of the language. This description is important in text translation because of loan words. The phonological analysis discusses the phonological processes involved in the adaptation of lexical borrowing from English to Akan and how the adaptation provides values that are closer to a native speaker’s perception and intuition. 

Akuapem Twi is an Akan dialect from the Niger-Congo Potou-Tano family cluster. It was the first Ghanaian language to achieve literary status with its transcription of the Bible. Spoken by the Akuapem ethnic subgrouping in the Eastern Region of Ghana and some parts of Cote d’Ivoire, the number of speakers  native to the language  is approximated to be 629,000 \citep{ethnologue}.  It shares mutual intelligibility with other Akan dialects, specifically Fante and Asante Twi. 

\begin{figure}
\centering
\includegraphics[width=8cm,height=3cm]{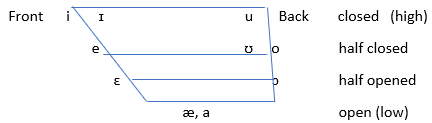}
\caption{Akuapem Vowel Chart}
\label{fig:vchart}
\end{figure}

Akuapem Twi is a tonal language with high, mid, and low tones which have contrastive semantic interpretations. It operates a 10 vowel system with feature specifications in oral, nasal, and ±ATR vowel harmony systems. The Akuapem vowel chart is shown in Figure \ref{fig:vchart}.

The different types of vowels are:
\begin{itemize}
    \item Oral vowels: /a, æ, i, \textsci, e, \textepsilon, o, \textupsilon, \textopeno, u/
    \item Nasalised vowels :  / \~i, ĩ, ã, \textscu, ũ /
    \item ±ATR Vowel Harmony 
\end{itemize}

The Advanced Tongue Root (ATR) is a regressive assimilation vowel harmony process in Akan \citep{Sefa}. In –ATR, the position of the tongue is pushed back whereas in the +ATR the tongue is pushed forward. \citep{Dolphyne} distinguishes the two groups as follows:
\begin{itemize}
    \item Advanced Tongue Root	 /i, e, æ, o, u/
    \item Unadvanced Tongue Root	/\textsci, \textepsilon, a, \textopeno, \textupsilon /
\end{itemize}
The +ATR and –ATR vowel harmony groups have a complementary distribution. The language forbids co-occurrence of the two sets within a word with more than a syllable \citep{APENTENGAMFO}.

The consonant inventory consists of the following:

\begin{itemize}
    
    \item /b, d, f, g, h, k, m, n, \textipa{\ng}, p, r, s, t/
   
    \item Glides :	/ w, j/
    \item Affricates :  [ \texttctclig, \textdzlig,\textctc,\textltailn, \textipa{\ng}, w,    h\textsuperscript{w}, k\textsuperscript{w}]
\end{itemize}
The language is characterised by extensive palatalisation. In the phonological phonotactics, when a front vowel follows the velar consonants /g, k/ or the glottal fricative /h/, a palatalised derivative is produced in the Surface Form.
\begin{itemize}
    \item /h, g, k/ $\rightarrow$ [\textctc,\textdzlig,\texttctclig]/ $\_$(+Front,-Low)
\end{itemize}
However there are exceptions to the rule where palatalisation does not occur in the set environment.
\begin{itemize}
    \item /h, k, g/ $\rightarrow$ [h, k, g]/$\_$(+Front,-Low)
\end{itemize}
Possible reasons for the exceptions and the alveo-palatals both existing in the  language have been attributed to amphichronic phonology and the present derivation admittedly being the product of the diachronic changes \citep{Adomako}.  This theory is proven in monolingual Akan speakers who forgo the palatalised phonemes in favour of the non-palatalised ones.

On syllabic directionality, Akuapem Twi operates to a large extent, a no coda syllable structure, and bans consonant clusters. The coda constraints in this directionality are sonorants/syllabic consonants which include nasals and glides (m, n,  ng, w). The syllabic consonants on their own constitute voiced syllables word final. Ergo, Akuapem Twi does not have a closed syllable structure. The syllable structures available in the language are: 
\begin{itemize}
    \item V     -    \hspace{2mm}\textbf{a}
    \item CV    -    a.\textbf{ba}           {seed} 
    \item CV.N  -  a.\textbf{ba.n}          {government} 
\end{itemize}
Just as the syllabic consonant is a syllable on its own, vowels in a sequence constitute syllables on their own and they carry tone as well \citep{Adomako,Dolphyne}. 

Though the language does not permit consonant clusters in its directionality, there appears to be an enduring violation to this constraint, the CCV, represented specifically as the CrV, (a type of syllable consisting of a consonant, the letter "r", and a vowel). Marfo and Yankson \citep{MarfoYankson} argue that the CCV, represented in Akan as the CrV, is the Surface Form (SF) of the Underlying Representation (UR) CV.rV, attributing it to a derivation that ‘has resulted from a phonological process that ends in a syllable reduction and a subsequent fusion of this reduced syllable into another’ \citep{Marfo}.

Phonological processes  like epenthesis, deletion and syllable structure modification are used to nativise foreign words that are borrowed into the native language.
In that regard, forms that exist in the source language do not necessarily have to undergo a lot of phonological processes to be adapted into the native language, case in point the CrV structure. See Table \ref{table:Loanwords1} for illustration.

    \begin{table} [t]
\centering
\begin{tabular}{ |c | c | c | }
\hline
 word   &   source language      &    native language \\
 \hline
 driver &   dra\textsci v\textschwa  &   dr\textopeno ba \\
\hline

\end{tabular}
\caption {Loanword adaptation for CrV syllable structure.}
\label {table:Loanwords1}
\end{table}
        
English operates both a closed and open syllable structure. When a word with a closed syllable structure is borrowed from English, the nativisation process would request that the word be phonologized as open in Akuapem Twi by way of an epenthetic vowel (insertion of a vowel). See Table \ref{table:Loanwords2} for an example.

      \begin{table} [t]
\centering
\begin{tabular}{ |c | c | c | }
\hline
 word   &   source language      &    native language \\
 \hline
 traffic &     traf\textsci k &   traf\textsci k\textbf{\textsci}  \\
\hline

\end{tabular}
\caption {Loanword adaptation for vowel insertion.}
\label {table:Loanwords2}
\end{table}

On the other hand, sounds in the source language that are not attested in the native language are replaced with an approximation of the sound in the native language.

The syntax of Akuapem Twi has a taxonomy of lexemes in both functional and lexical categories. Major lexical categories are nouns, verbs and adjectives. 
 
Nouns can be subjects or objects of a predicate. They can also take on adjuncts such as adjectives and determiners and be morphologically inflected with phi-features – gender, number. 

Verbs indicate tense, aspect and negation with morphophonological inflections and are always placed between the subject and the object except when they are in the imperative.
Usually, the root of the verb form which is the simple present tense is inflected with a  set of functional affixes to indicate different tenses. An illustration is presented in Table \ref{table:Tense}. 

\begin{table} [t]
\centering
\begin{tabular}{ |c | c | c | c |}
\hline
 di   &   eat         &    SIMPLE PRESENT \\

\hline
 re-di  &   is eating   &   PROGRESSIVE  \\
\hline
be-di     &     will eat     &       FUTURE \\
\hline
     a-di  &   has eaten    &  PRESENT PERFECT \\
 \hline
         di-i  &   ate         &   PAST \\
 \hline

\end{tabular}
\caption {Akuapem Twi Indicating Tense}
\label {table:Tense}
\end{table}

Both the nominal subjects and objects can have adjuncts such as adjectives and determiners together to form a constituency. The determiners mostly follow the noun as well but can precede it in a few instances.
 The word order of Akuapem Twi is the nuclear predication, SVO – Subject  Verb Object. For example:
 
[[Papa no] (S)\hspace{2mm} [nnoaa] (V) \hspace{3mm}[hwee] (O) \hspace{6mm}bio]

 \hspace{2mm}\textit{Man the\hspace{6mm} has not cooked anything \hspace{6mm}again.}

 \textbf{The man has not cooked anything again}
 
Questions in the language are raised in two ways: tone and the interrogative marker `so'. In tone, a declarative sentence can become a question by raising the tone at the end of the sentence. This is illustrated as follows.

\begin{enumerate}

\item Papa no nnoaa hwee bio?
 
\item So papa no nnoaa hwee bio?
\end{enumerate}

\section{Parallel Corpora}
Parallel corpora or translation corpora involve texts and their translations. Source and target texts may be aligned at word, sentence or paragraph level\citep{zanettin2014translation}.
According to Li \citep{doval2019parallel}, the first parallel corpora was the Canadian Hansard Corpus which consisted of Canadian parliamentary proceedings published in English and French. Other parallel corpora mostly available online are Multilingwis, EPTIC ( European Parliament Translation and Interpreting Corpus), Cluvi and Intercorp. These resources contain authentic examples of previously translated texts unlike online automatic aids like Google Translate, Bab.la, Babelfish and Systran.
Much like these parallel corpora, the unidirectional English- Akuapem Twi parallel corpus for this study may serve as resource for research in many domains including Natural Language Processsing (NLP), computational linguistics and lexicography.
Furthermore, since corpora are usually open-ended, the unidirectional English- Akuapem Twi corpus may be enlarged and uploaded online to develop general language tools and computer- assisted translation tools such as electronic dictionaries, spell- checker programmes, translation  memories (TMs), concordances and terminology banks \citep{genette2016reliable}.  As more tools are developed, more authentic translations can be done to overcome the challenge of scarcity of electronic texts in a low resource dialect of Akan like Akuapem as well as other Ghanaian languages and to speed up the evolution of  Akan from simple online existence to optimal online presence.

\subsection{Existing Datasets in Akan}
The amount of multilingual text available for Twi is not as extensive as it should be, and this section seeks to explore the current state of such data. For languages with greater resources, there is a significant amount of text available online for the use of NLP, with some of the best sources being Wikipedia and the Bible. For lesser-resourced languages, the Bible is the most available resource \citep{Resnik}. It is essential to consider that many of these resources typically have a non-native and religious bias, resulting in low translation quality and noisy datasets. Recently, \citet{YorubaandTwipaper} explored the major existing Twi datasets -- Bible translation pairs (labeled as clean), Jehovah’s Witness data, Wikipedia data, and the JW300 Twi corpus \citep{agic2020jw300}. The latter 3 are assessed and labeled as noisy due to dialectal inconsistencies. These inconsistencies make it crucial to create cleaner data. 

The JW300 Twi corpus and Jehovah’s Witness datasets are significant in size, together providing over 1.5 million word tokens. However, it is important to be aware of the religious influences of this data and the errors it contains, making this corpus noisy for NLP tasks. The JW300 dataset presents the corpus in an incredibly convenient way, available as parallel translation sentence pairs of English to Akuapem Twi, though there is a slight mixing of the Akuapem and Asante dialects within this data, contributing to the noise. \citet{YorubaandTwipaper} further explored the ways that NMT tasks can be supported within a small dataset of just clean data. Utilizing the clean data provides a higher accuracy assessment from native speakers, but as the dataset grows, the use of noisy data becomes necessary and leads to greater errors.

Another religiously biased dataset is the Bible, explored for NLP purposes by \citet{YorubaandTwipaper} and \citet{BibleCorp}. This dataset contains 600,000 tokens and is the cleanest dataset currently available in Twi. This Bible data is also presented as a parallel English to Asante Twi corpus. This parallel corpus was additionally used by \citet{BibleCorp} to conduct neural machine translation with state-of-the-art MT architectures. Within the research conducted by \citet{BibleCorp}, the Twi corpora contained a single version, whereas the English consisted of four different versions: King James Version, Good News Bible, Easy to Read Version, and New International Version. 

The TypeCraft (TC) Akan Corpus was created to improve the development of online lexicographical data for lesser resourced languages \citep{TCAkan}. The TC-Akan Corpus contains 261 pieces of text, with 98,000 tokens. These words were annotated for part-of-speech and glossed for the TC-Akan Corpus, leading to 1,367 words within this dictionary. This project was spurred by the existence of two physical Twi dictionaries, which are fairly comprehensive, but are not in a format that is supported online. Another important feature of the TC-Akan corpus is its refinement of current online dictionaries, available from the Ghana Institute of Linguistics Literacy and Bible Translation (GILLBT). The compilation of words from these dictionaries had to be refined for dialectal and spelling accuracy, coupled with the removal English loan words to ensure a clean corpus.

Considering the synthesis of all of this data and the growing need for neural MT in Twi, creating data that is clean and widely available is necessary.

\section{Methodology}
In this section we describe the methodology for generating our parallel corpus.

\subsection{Source English Sentence Data}
The corpus of English sentences used as the source text was
curated from tatoeba.org \citep{Tatoeba.org} -- this data is licensed under the Creative Commons - Attribution
2.0 France license (terms of use on page). The sentences originally appeared as
part of a bilingual corpus for English and German. A total of 50,000 English sentences were randomly sampled from the above dataset and used for our project.

\subsection{Crowd-sourced Parallel Data}
The crowd-sourced data was obtained from voluntary responses of people through a Google
Form survey over a period of 2 months. A set of English sentences were randomly selected from the source English sentence data for volunteers to respond
by providing the correct Twi translations. The responses were manually verified and corrected by in-house native speakers and linguists. This often involved correcting wrongly used special characters. 
However, it is important to mention that these crowd-sourced parallel data were not scored during the verification phase as compared to our machine generated preliminary translations.

\begin{table*} [t]
\centering
\begin{tabular}{ |c || c | c | c |}
\hline
\multicolumn{4}{|c|}{Human Verified Corpus} \\ [1ex]
    \hline
     & sentence count & word count & unique word count \\
    \hline
    English & 25,421 & 163,861 & 13,058 \\
    \hline
    Akuapem Twi & 25,421 & 170,908 & 12,314 \\
    \hline
\multicolumn{4}{|c|}{Crowd-sourced Data Corpus} \\ [1ex]    \hline
     & sentence count & word count & unique word count\\
    \hline
    English & 697 & 3,525 & 1,225 \\
    \hline
    Akuapem Twi & 697 & 6,702 & 2,541 \\
    \hline   

\end{tabular}
\caption {Dataset Statistics}
\label {table:Metadata}
\end{table*}

\subsection{Distribution of sentences to researchers}
As part of the process of generating 50,000 English to Akuapem Twi sentence pairs, we used the transformers library by Hugging Face \citep{Huggingface} to load the aforementioned OPUS-MT pretrained model from the Language Technology
Research Group at the University of Helsinki \citep{TiedemannThottingal:EAMT2020}. We first tuned some hyperparameters of the model in order to improve the preliminary translations. We found the temperature and beam search settings to be the most important. Specifically, we used a temperature setting of 1.0, with 4 beams and early stopping enabled for beam search.

Ten (10) researchers from NLP Ghana had earlier been nominated, based on their knowledge and fluency of the Akuapem Twi language, to work as members of the
team for scoring, verification and correction of the preliminary translations. As we had initially aimed to translate 50,000 sentences, 10 researchers were provided with 5,000 sentence pairs each for revision. As a form of motivation or compensation for their work, every researcher was offered \$1 for 50 scoring, verification and corrections.
The 5,000 sentence pairs were saved in a google sheet and distributed to each researcher.

\subsection{Verification and Correction of Preliminary Translations}

The typical process of verification of translations involved a researcher who is also a native speaker of the
local language.  In other words, researchers reviewed the preliminary translations generated by the translator. Based on the
accuracy of the translation as determined by the researchers who were human translators, these preliminary translations were either
maintained or revised.

\subsection{Final Dataset}

At the initial stages of the translation process, we learned that the professional linguist market rate
is significantly higher at ~1\$ per 1 translation on average. Perhaps unsurprisingly, we were
unable to hit 100\% of our target, but achieved a decent 50+\% completion within the agreed
timeline of two months, as presented next.

\subsubsection{Statistics of Human-verified Corpus}
Overall, due to monetary and time constraints approximately 25,000 of English to Akuapem Twi sentence pairs were scored, verified and corrected where necessary by 10 native speakers. Final statistics of the dataset are described in Table \ref{table:Metadata}.

\subsection{Evaluation}
The crowd-sourced data corpus was already translated by native speakers hence was already verified. The main evaluation for the dataset was therefore on the 25,000 English to Akuapem Twi sentence pairs. 

Ten (10) native Twi speakers were employed to verify the preliminary translations from the model. This is similar to the use of humans for verification employed by \citep{chen2018emotionlines} in verifying the emotions associated with a particular text. Instead of five people checking a single sentence as used in \citep{chen2018emotionlines}, one person was employed to check on the quality and clarity of a specific translated sentence. They would  then score it on a scale of 1 to 10, with 10 being a perfect translation. A scoring threshold of eight (8) was established, below which the translated sentence was edited to make more sense, and above which it was left as-is. At a score of 8, another person double-checked the quality of the translated text to see if it needed editing.

\section{Translator Fine-Tuning}
After the human-verified data was prepared, it was used to fine-tune the aforementioned OPUS-MT transformer model further. The weights of the transformer architecture, implemented with the transformers Python library by Hugging Face (with PyTorch backend), were initialized to the Helsinki OPUS-MT model. All the model layers were then fine-tuned using the Adam optimizer \citep{kingma2014adam} for 20 epochs. This was done on a single Azure Cloud GPU-enabled NC6 Virtual Machine, with a batch size of 8. This took 2 hours and 20 minutes to execute, achieving a loss value of about 0.12.

The well-known Bilingual Evaluation Understudy (BLEU) \citep{papineni2002bleu} score for measuring the quality of text which has been translated from our algorithms is the metric used to measure the quality of the translations. In BLEU, there is a “reference” -- a human translated version of a sentence, as well as a “candidate” -- translation generated by the algorithm. The “candidate” is compared with the “reference” and the BLEU score, which is a modified precision measure, is generated \citep{papineni2002bleu}. While human evaluation of machine translations is expensive, BLUE is fast, less expensive and language-independent. Moreover, it has a high correlation with evaluations from humans \citep{papineni2002bleu} though the degree of high correlation differs by language-pairs as seen with submissions on the WMT metrics shared task \citep{WMT}.  

On our evaluation of the model, BLEU scores on the “candidate” sentences (translated sentences from the fine-tuned model - from English to Akuapem Twi) are generated using the \textit{corpus-bleu} function in the Python \textit{nltk bleu-score} module. The test data used is the crowd-sourced data of 697 English sentences and 697 Akuapem Twi “references”. The highest BLEU score achieved was 0.720 using the following parameters: \textit{smoothing function} value of \textit{7}, with \textit{auto-reweigh} of \textit{True} and \textit{weight} tuple values of \textit{(0.58,0,0,0)}, indicating the focus is on “adequacy” instead of “fluency” in the translations \citep{papineni2002bleu}. In comparison, the OPUS-MT model before fine-tuning scored 0.694 using the same parameters indicating the fine-tuned model made a noticeable improvement of 3.75\% over the OPUS-Model. Subjectively, the fine-tuned model was reported as a smoother experience by testers, who found it to yield egregiously wrong translations less frequently.

\section{Result and Discussion}
We found translations of the crowd-sourced and verified data to be largely accurate and acceptable. Many of the sentences in the source text are simple and express universal concepts which are rendered very close in the target language, while taking into consideration the target culture and conventions. Akuapem Twi is very close to Asante Twi, and many of the translations are correct in both dialects of Akan.

A number of translation strategies were used depending on the translation difficulty encountered. These strategies include loaning (borrowing), equivalence, cultural substitution, translation by a more general word among others. For example, words like \textit{Paris, Boston, Australia} and \textit{camera} were borrowed completely from the source language. Nonetheless, others like \textit{coffee, computer} and \textit{gas} were borrowed and adapted in terms of orthography and phonology -- as \textit{k\textopeno fe}, \textit{k\textopeno mputa} and \textit{gaas} respectively. There is also an attempt to provide several translations of the same source item in different contexts. This is useful for words like \textit{welcome} which may mean \textit{akwaaba} or \textit{ndaase nni h\textopeno}  depending on the context.

However, a closer look at the translations reveals the need for further improvement to the presented corpus. The difficult and time-consuming process of compiling corpora especially in African languages like Akan -- which is yet to emerge from simple online existence to optimal online presence -- may account for a number of errors in the crowd-sourced translations. 
In this regard, since one of the aims of the parallel corpus is to facilitate further training of machine translation models in Akuapem Twi, and considering that corpora are usually open-ended, the translations must be revised further and updated as well to enlarge and improve the quality. For example, we recently identified what appears to be a mistranslation in entry 86 of the human-verified data, where \textit{turn on the gas} is rendered as \textit{fa w’ani si gaas no so} -- \textit{keep an eye on the gas}. A more appropriate translation might be \textit{s\textopeno}  \textit{gaas no}.

There is also the need to pay attention to, for example, cultural sensitivity by providing alternative translations to taboo words in Akuapem Twi and Akan in general to enrich the corpus. It is noteworthy that natives often prefer euphemistic terms, and these may even be found in texts within specialised fields like Medicine. The human translator might need to consider this, together with other aspects of culture for the target audience, when improving the corpus in the future.

Additionally, machine translations generated by the fine-tuned transformer model are mostly grammatically correct, but sometimes not idiomatic. Human touch or post-editing remains crucial to ensure higher accuracy and quality.
Other examples underlining the need for the human touch after machine translation are presented in Table \ref{table:HumanTouch}.

\begin{table*} [t]
\centering
\begin{tabular}{ |c | c | c | c |}
\hline
   SOURCE TEXT & OPUS MT TRANSLATION &  CORRECTION \\

\hline
  I've gotten better   &   M'ani agye yiye  &  Meho at\textopeno me \\
\hline
 How did it go last night?  & \scalebox{1.4}{\textepsilon}y\textepsilon \textepsilon \hspace{0.3mm} d\textepsilon n  na ade kyee    &       Nn\textepsilon ra anadwo \textepsilon kosii s\textepsilon n?\\
\hline

\end{tabular}
\caption {Human Touch After Machine Translation}
\label {table:HumanTouch}
\end{table*}

In the two examples above, the OPUS MT Translation of both conform to punctuation rules syntax of Akan that has a subject-verb- object (SVO) structure. However, both translations are misleading in terms of meaning and require revision by a human translator as indicated in the correction/OPUS MT Translation after Fine-tuning column.
Finally, from our previous experiences we hope to offer our human translators compensations commensurate with the tasks we give them in the future. This will ensure that they score, verify and correct machine generated translations with minimal errors.

\section{Conclusion}
In this paper, we have presented a bilingual machine translation training corpus of 25,421 sentence pairs for English and Akuapem Twi. We used a transformer-based translator to generate initial translations into Akuapem Twi, which were later verified and revised where necessary by native speakers. The main idea of a typical use case for the dataset is for further training of machine translation models in Akuapem Twi. The data can also be used for other downstream NLP tasks such as named entity recognition and part-of-speech tagging, with appropriate additional annotations. Another potential application is fine-tuning an unsupervised embedding for the Akuapem Twi dialect of the Akan language.

In addition, about 697 crowd-sourced sentences of a higher quality are provided for use as an
evaluation set for the tasks highlighted above. It is recommended as a testing dataset for English to Twi and Twi to English machine translation models.

In  order to contribute towards the growth of the African NLP community, especially in the area of research and development, the dataset is \href{https://zenodo.org/record/4432117#.YF5rndKSk2y}{publicly available}.

\section*{Acknowledgments}
We are grateful to Microsoft for Startups Social Impact Program -- for supporting this research effort via providing GPU compute through Algorine Research. We would also like to thank Julia Kreutzer and Jade Abbot for their constructive feedback. This project was also supported by the \href{https://www.k4all.org/project/language-dataset-fellowship/}{AI4D language dataset fellowship} through K4All and Zindi Africa. We would also like to thank the \href{https://gajreport.com}{Ghanaian American Journal} for their work in sharing our work and mission with the Ghanaian public.

\bibliography{anthology,eacl2021}
\bibliographystyle{acl_natbib}

\appendix

\end{document}